%% file: sample-sigconf.tex
\documentclass[acmtog]{acmart}

\usepackage{booktabs}
\usepackage{tabularx}
\usepackage{array}
\usepackage{graphicx}
\usepackage{multirow}
\usepackage{enumitem}
\usepackage{microtype}
\usepackage{colortbl}
\AtBeginDocument{%
  }

\setcopyright{none}
\renewcommand\footnotetextcopyrightpermission[1]{}

\title[Video Models as Native 4D Renderers]{Video Models as Native 4D Renderers:\\
World-Grounded Conditioning from Animated Mesh}

\author{Junhao Chen}
\email{yisuanwang@gmail.com}
\affiliation{%
  \institution{Tsinghua University}
  \city{Shenzhen}
  \country{China}
}

\author{Mingjin Chen}
\email{chenmingjin1238@gmail.com}
\affiliation{%
  \institution{The Hong Kong Polytechnic University}
  \city{Hong Kong}
  \country{China}
}

\author{Henghaofan Zhang}
\email{hhfzhang@outlook.com}
\affiliation{%
  \institution{University of Electronic Science and Technology of China}
  \city{Chengdu}
  \country{China}
}

\author{Minglin Chen}
\email{chenmlin8@mail2.sysu.edu.cn}
\affiliation{%
  \institution{Sun Yat-sen University}
  \city{Guangzhou}
  \country{China}
}

\author{Liaoyuan Fan}
\email{u3619617@connect.hku.hk}
\affiliation{%
  \institution{The University of Hong Kong}
  \city{Hong Kong}
  \country{China}
}

\author{Boran Zhang}
\email{chenqingsui1@gmail.com}
\affiliation{%
  \institution{University of Science and Technology of China}
  \city{Hefei}
  \country{China}
}

\author{Saining Zhang}
\email{saining002@e.ntu.edu.sg}
\affiliation{%
  \institution{Nanyang Technological University}
  \city{Singapore}
  \country{Singapore}
}

\author{Mingze Sun}
\email{smz22@mails.tsinghua.edu.cn}
\affiliation{%
  \institution{Tsinghua University}
  \city{Beijing}
  \country{China}
}

\author{Hao Zhao}
\email{zhaohao@air.tsinghua.edu.cn}
\affiliation{%
  \institution{Tsinghua University}
  \city{Beijing}
  \country{China}
}

\author{Ruqi Huang}
\authornote{Corresponding authors: Ruqi Huang and Yufei Wang.}
\email{ruqihuang@sz.tsinghua.edu.cn}
\affiliation{%
  \institution{Tsinghua University}
  \city{Shenzhen}
  \country{China}
}

\author{Zhihao Li}
\email{zhihao.li@sparclab.ai}
\affiliation{%
  \institution{SparcAI Inc.}
  \streetaddress{221 W 9th St PMB 141}
  \city{Wilmington}
  \state{DE}
  \postcode{19801}
  \country{USA}
}

\author{Yufei Wang}
\authornote{Corresponding Author.}
\email{yufei.wang@sparclab.ai}
\affiliation{%
  \institution{SparcAI Inc.}
  \streetaddress{221 W 9th St PMB 141}
  \city{Wilmington}
  \state{DE}
  \postcode{19801}
  \country{USA}
}

\begin{teaserfigure}
  \centering
  \makebox[\textwidth][c]{\includegraphics[width=\textwidth]{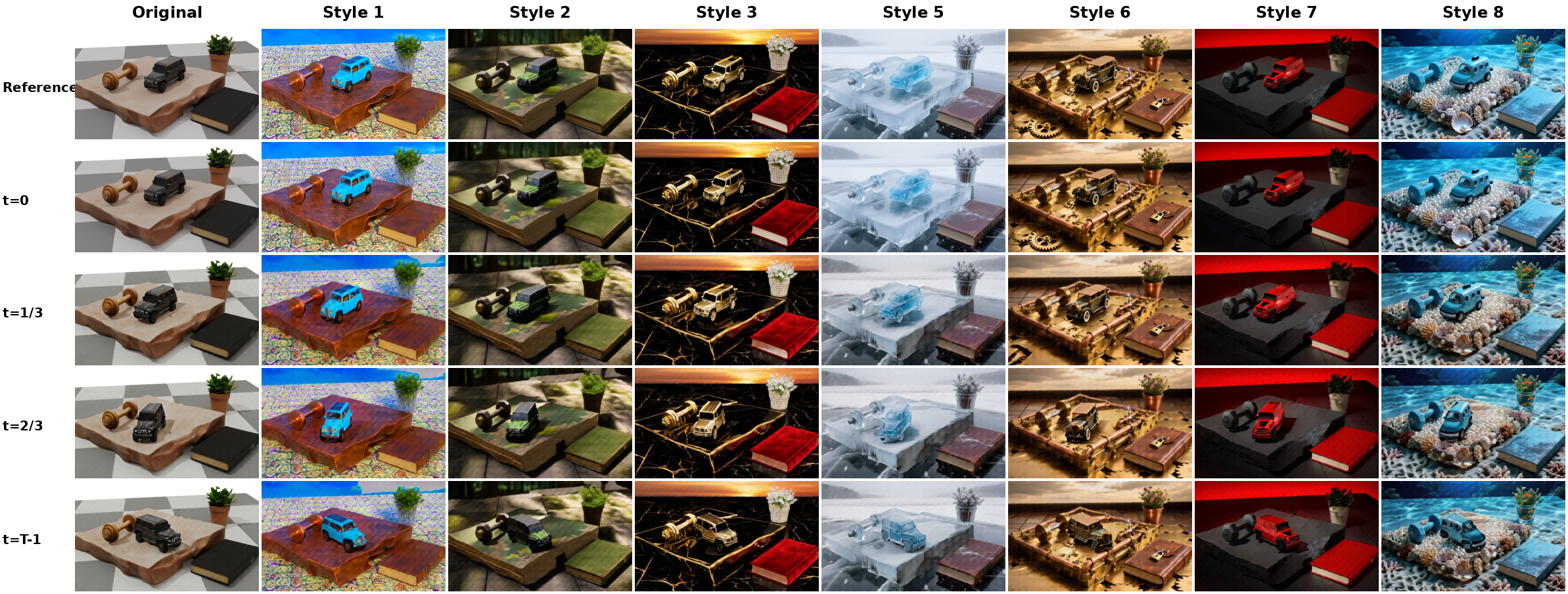}}
  \caption{\textbf{DAR is a reference-guided 4D renderer: with the animated mesh and camera fixed, changing only the first-frame reference image re-renders the scene's appearance.} The same animated car mesh, camera trajectory, and 4D geometry are held fixed; each column uses a different reference style and each row is a normalized animation time.  DAR alters paint style, color, and vehicle appearance while keeping the camera, pose, and silhouette aligned, indicating partial decoupling between geometry and appearance.}
  \Description{Teaser figure showing reference-image appearance variants for a fixed animated car mesh and camera trajectory; each column uses a different first-frame reference, producing different appearance styles while preserving pose and silhouette.}
  \label{fig:ref_variants}
\end{teaserfigure}

\begin{abstract}
\input{tog26_sections/00_abstract}
\end{abstract}



\begin{document}

\fancypagestyle{firstpagestyle}{%
  \fancyhf{}
  \renewcommand{\headrulewidth}{0pt}
  \fancyfoot[C]{\thepage}
}
\pagestyle{plain}

\maketitle

\section{Introduction}
\input{tog26_sections/01_introduction}

\section{Related Work}
\input{tog26_sections/02_related_work}

\section{Method}
\input{tog26_sections/03_method}

\section{DAR-4D Dataset}
\input{tog26_sections/04a_dataset}

\section{Experimental Setup}
\input{tog26_sections/04_setup}

\section{Results}
\input{tog26_sections/05_results}

\section{Conclusion}
\input{tog26_sections/07_conclusion}

\clearpage
\bibliographystyle{ACM-Reference-Format}
\bibliography{tog26_references}

\clearpage
\onecolumn

\begin{figure}[p]
  \centering
  \vspace*{-1.0em}
  \makebox[\textwidth][c]{\includegraphics[width=1\textwidth]{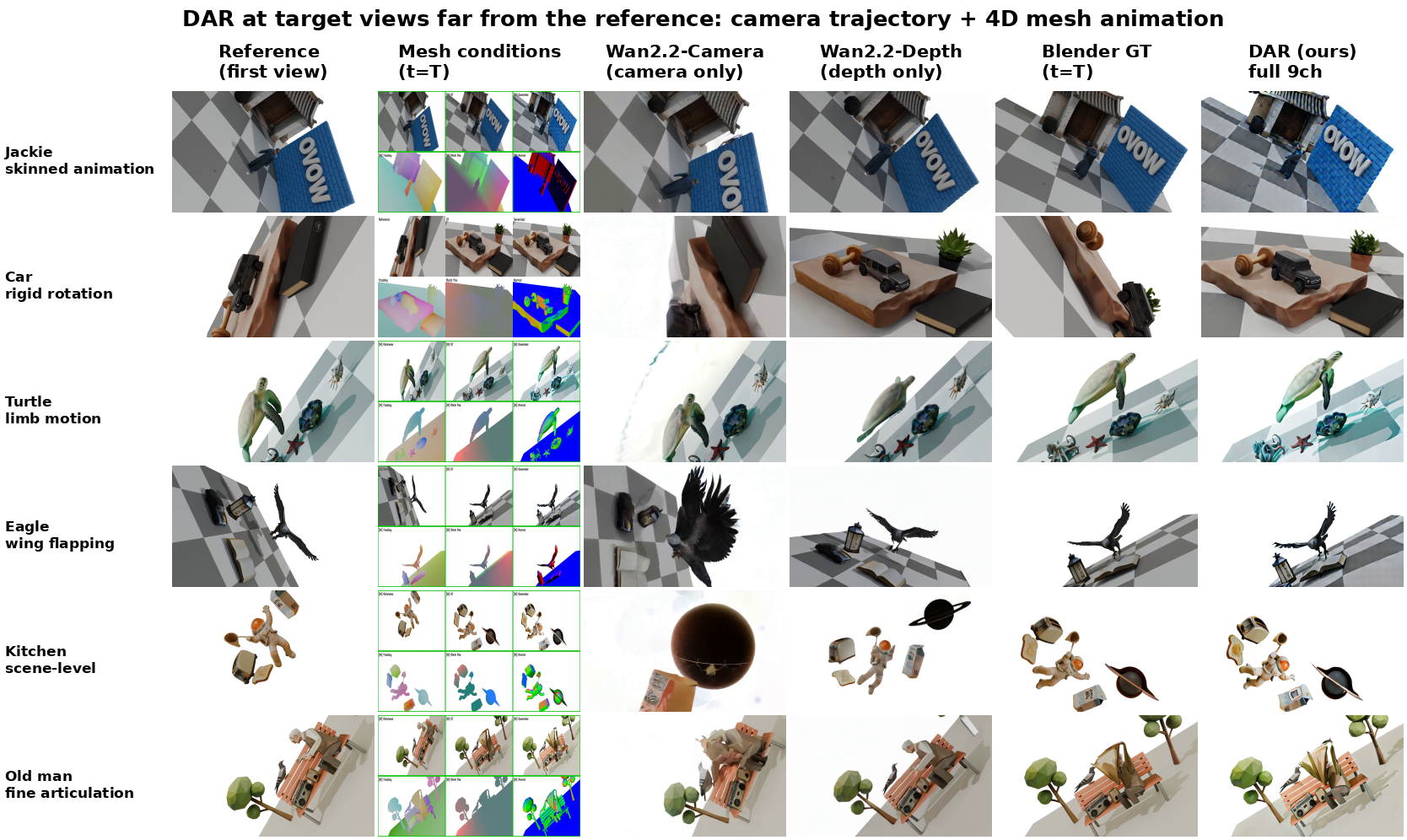}}
  \caption{\textbf{DAR controls both camera trajectory and in-scene 4D mesh animation for generative rendering.} Each row shows one OVOW Blender asset under a $\pm60^{\circ}$ horizontal orbit.  Except for the reference column, all panels show the final frame $t{=}T$, where the target view is farthest from the first-frame reference.  Columns show reference, mesh-derived 9-channel geometry, Wan2.2-Camera, Wan2.2-Depth, Blender GT, and DAR.  Wan2.2-Camera lacks per-frame object geometry and hallucinates the foreground; Wan2.2-Depth couples camera and object motion; DAR jointly injects Pl\"ucker camera rays and 9-channel geometry to better match both the target camera and mesh animation.}
  \Description{Comparison figure showing DAR vs Wan2.2-Camera and Wan2.2-Depth on six OVOW animated Blender scenes under a shared orbit camera trajectory; DAR better respects both camera path and 4D object animation.}
  \label{fig:main_qualitative}
\end{figure}

\clearpage

\begin{figure}[p]
  \centering
  \vspace*{-1.0em}
  \makebox[\textwidth][c]{\includegraphics[width=1\textwidth]{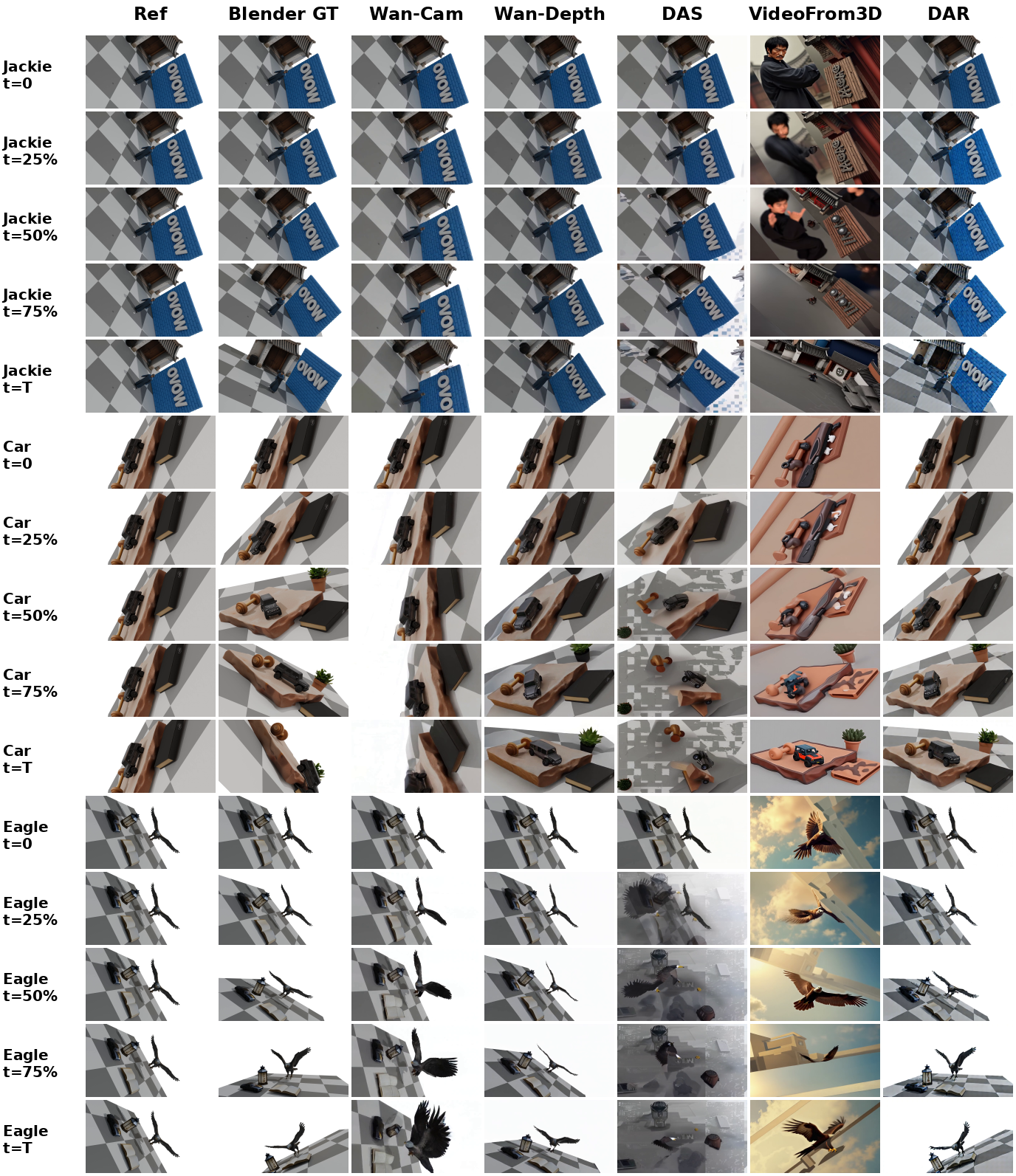}}
  \caption{\textbf{Executed baseline suite across animation time.} Three OVOW assets are shown at five normalized time points ($t{=}0/25\%/50\%/75\%/T$) across Reference, Blender GT, Wan2.2-Camera, Wan2.2-Depth, DAS, VideoFrom3D, and DAR.  All videos are sampled by normalized animation time.  DAR more consistently preserves both target camera trajectory and animated mesh state.}
  \Description{External baseline suite comparing Wan2.2-Camera, Wan2.2-Depth, DAS, VideoFrom3D, and DAR on OVOW animated Blender scenes.}
  \label{fig:baseline_suite}
\end{figure}

\clearpage
\twocolumn
\appendix
\input{tog26_sections/08_appendix}

\end{document}

%% file: tog26_sections/00_abstract.tex
Pretrained video diffusion models can act as renderers when the desired scene state is already specified by an animated mesh, a camera trajectory, and a reference image.  This \emph{4D generative rendering} setting raises a representation question: what image-format condition lets a video backbone obey both camera motion and scene-internal animation?  We propose DAR, a reference-guided renderer that extends Wan2.2 camera control from Pl\"ucker rays alone to a joint camera-plus-geometry interface.  DAR projects a neural 4D G-buffer (tracking, world position, and normal) from the animated mesh and injects it through a widened control adapter while preserving the pretrained image-to-video prior.
The central design choice is the pair of tracking and world position.  Tracking identifies the persistent surface element that should carry appearance; world position gives its current scene-coordinate state; normal supplies local shape.  Depth plus calibrated rays can recover 3D in principle, but depth is a camera-dependent chart in which camera and object motion are mixed.  On the 68-case DAR-4D benchmark, LoRA DAR reaches PSNR 23.22, SSIM 0.895, and LPIPS 0.134, improving over off-the-shelf Wan2.2-Depth by 1.54 dB PSNR; a full fine-tune reaches PSNR 25.36 and SSIM 0.917.  Matched ablations show that replacing world position by depth reduces PSNR by 1.26--1.55 dB at every checkpoint, supporting tracking+world-position correspondence as a practical 4D rendering condition.

%% file: tog26_sections/01_introduction.tex
Generative video models are increasingly useful as rendering engines.  In many graphics workflows, a user or simulator already specifies the 4D state of a scene: an animated mesh, a camera path, and a reference appearance.  The target is therefore not an unconstrained video, but a rendered video that preserves the specified camera trajectory and scene motion while adding material, lighting, texture, and high-frequency detail.  This problem appears in previz, look development, game cinematics, synthetic-data generation, embodied simulation, and world-model visualization.  We call it \textbf{4D generative rendering}.  The difficulty is that ordinary videos entangle observer motion and object motion, while a renderer must keep them separately controllable.

Existing controllable video models expose only part of this interface.  Camera-control models such as Wan2.2-Camera follow a target trajectory through Pl\"ucker rays, but they receive no per-frame object geometry, so foreground identity and animation are left to the image-to-video prior or a text prompt~\citep{wang2025wan}.  Depth-conditioned models receive a strong projected layout signal, but depth is measured in the current camera frame and does not identify which moving surface point is being observed~\citep{wang2025wan,bytedance2025seedance,alhaija2025cosmos}.  Tracking-guided and 3D-assisted methods move closer to generative rendering, but sparse or object-centric tracks weaken under novel views and occlusion, while static-scene anchor methods assume the surface state does not change over time~\citep{gu2025diffusionshader,kim2025videofrom3d,cai2024generativerendering}.  The common limitation is representation: a single camera map, depth map, prompt, or sparse tracking image cannot tell the model both which ray is being rendered and which animated surface state lies on that ray.

We propose \textbf{DAR (Diffusion as Renderer)}, a reference-guided video renderer built around a \textbf{neural 4D G-buffer} that uses a pretrained video diffusion model as the renderer.  As shown in Fig.~\ref{fig:pipeline}, the animated mesh is rasterized into tracking, world-position, and normal maps; the target camera is represented by per-pixel Pl\"ucker rays; and a first-frame reference image supplies appearance.  Architecturally, we keep the Wan2.2-Fun-5B-Control-Camera backbone and make one targeted change: the original camera-control adapter is widened from 24 Pl\"ucker channels to $(24{+}N)$ camera-plus-geometry channels, then fine-tuned with LoRA.  The transformer therefore receives, at the same early control point, the ray being rendered, the animated surface state on that ray, and the reference appearance to propagate.

The key condition is \textbf{tracking plus world position}.  Tracking tells the model which persistent surface element should carry appearance; world position tells where that element is in the current 3D scene.  Unlike depth, this pair separates surface identity from camera measurement and keeps observer motion in the Pl\"ucker channels.  This follows a long graphics and vision principle: view-consistent synthesis is easier when a model is given dense coordinate or correspondence maps, as in functional maps, DensePose, NOCS, UV position maps, and UV-space texture diffusion~\citep{ovsjanikov2012functional,guler2018densepose,wang2019nocs,feng2018prnet,yu2023pointuv,zeng2024paint3d,chen2026ultraman}.  DAR brings this principle to dynamic mesh-to-video rendering by using tracking+world position as a visible 4D state code.

We evaluate DAR on \textbf{DAR-4D}, a multi-source rendering corpus and synthesis pipeline built from rigged animations, physics scenes, public/third-party 3D assets, BlenderKit scenes, and Unreal Engine environments.  On the 68-case \texttt{4d\_vis} benchmark, LoRA DAR reaches PSNR \textbf{23.22}, SSIM \textbf{0.895}, and LPIPS \textbf{0.134}, improving over off-the-shelf Wan2.2-Depth by \textbf{+1.54 dB} PSNR.  A full fine-tune of the same condition reaches PSNR \textbf{25.36} and SSIM \textbf{0.917}.  In the matched depth-swap ablation, where only world position is replaced by depth, world position improves PSNR by \textbf{+1.42 dB} at checkpoint 10k and by 1.26--1.55 dB across all saved checkpoints.  Qualitative comparisons with Wan2.2-Camera, Wan2.2-Depth, DAS, and VideoFrom3D show that camera-only, depth-only, and sparse/static 3D controls each fail on one axis of the renderer interface.

Our contributions are:
\begin{itemize}
  \item a formulation of \textbf{4D generative rendering} as reference-guided video rendering from an animated mesh and target camera path;
  \item a \textbf{neural 4D G-buffer} condition that pairs Pl\"ucker rays with tracking, world position, and normal maps;
  \item a geometric and empirical argument that \textbf{tracking+world position} is a better visible 4D state code than camera-dependent depth for dual camera/object control;
  \item \textbf{DAR-4D}, a renderer-ready corpus and benchmark. It contains RGB videos, calibrated cameras, Pl\"ucker rays, and per-frame buffers for depth, normal, world position, and tracking.
\end{itemize}

%% file: tog26_sections/02_related_work.tex
\label{sec:related}

\subsection{Animation Production Pipelines}
Bringing an animated shot to the screen has traditionally meant committing to a full production pipeline, whose form depends on the medium.  \emph{2D animation} composes hand-drawn or vector keyframes that are inbetweened and composited, and recent methods learn to synthesize such vector animations directly~\citep{Chen_2026_CVPR_LottieGPT}.  \emph{3D and CG animation} for film and games instead builds an explicit world: assets are modeled, generated from multimodal or compositional inputs~\citep{chen2025idea23d,weng2026garmentgpt}, and edited~\citep{weng2026feedforward3deditinglearns}, then rigged and driven by learned skeleton and motion generation~\citep{sun2025drive,Sun_2026_CVPR_Animator}, and finally textured, lit, and rendered with extensive support from texture and material synthesis and physically based (neural) rendering~\citep{chen2023text2tex,gao2024genesistex,zeng2024paint3d,liang2025diffusionrenderer,xue2025pbrinspired}.  \emph{Scene-level animation} scales this to whole environments that must stay consistent across views and over time~\citep{chen2024scenetex,huang2025roompainter,hollein2024viewdiff}, while \emph{video generation} animates directly in pixel space through pose- and skeleton-conditioned character synthesis~\citep{xu2024magicanimate,chen2026dancetogether} and general controllable generators~\citep{wang2025wan,jiang2025vace}.  These pipelines reach high visual quality, but they sit at two extremes: classical CG needs a fully specified renderable scene, whereas pure video generation leaves the underlying 3D/4D state implicit and hard to control precisely.

\subsection{Generative Rendering with Video Diffusion}
This gap motivates \emph{generative rendering}: rather than authoring materials and lighting and invoking a classical renderer, one specifies only the scene geometry and lets a pretrained video diffusion model act as the renderer~\citep{cai2024generativerendering,gu2025diffusionshader,kim2025videofrom3d,huang2026generativeworldrenderer,zhang2025i2v3d}.  The open question is which geometric condition to expose.  Camera-control models steer viewpoint through ray or pose signals~\citep{wang2025wan,bai2025recammaster}, depth-, edge-, and normal-conditioned systems use projected geometry for layout~\citep{bytedance2025seedance,alhaija2025cosmos}, and 3D-conditional or tracking-guided methods render from coarse geometry, tracked points, or explicit 3D inputs~\citep{Chen_2026_CVPR_hvg3d}.  Their control is typically single-axis, however: camera controls ignore the animated surface state, depth is view-dependent, and static anchors do not handle articulated motion.  A parallel lineage shows that dense coordinate maps ease cross-view reasoning, from functional maps, DensePose, and NOCS to UV-space diffusion and neural G-buffers~\citep{ovsjanikov2012functional,guler2018densepose,wang2019nocs,feng2018prnet,miao2026framessequencestemporallyconsistent,yu2023pointuv,chen2024unirenderer}.  DAR unifies these views: it keeps Wan2.2's camera prior but conditions camera rays and a per-frame 4D G-buffer together, giving one control signal for both observer and object motion.

\subsection{Dynamic 4D Data and Synthetic Scene Generation}
Training and evaluating a video renderer requires paired dynamic data, and a growing set of resources supplies parts of it.  Large static-object and scene collections provide geometry and appearance diversity~\citep{deitke2023objaversexl,dl3dv}, synthetic engines and generative domain randomization render controllable data with ground-truth buffers~\citep{greff2022kubric,geng2025viewworldssingleimage3d}, and non-rigid, dynamic-stereo, and 4D or physics-oriented corpora add deformable and time-varying supervision~\citep{li2021deformingthings4d,karaev2023dynamicstereo,dynamicverse,physinone2026,wu2025cat4d}.  A recent line turns monocular or multi-view video into 4D or world-scale scenes~\citep{chen2026videoworldturningmonocular,chen2026enginenativeeditable3dworld,chen2025deepverse,yang2026neoverse}, which is exactly the kind of pipeline that yields paired video and 4D-scene data.  DAR-4D is complementary to these resources: rather than proposing new capture or reconstruction, it converts heterogeneous assets into a single renderer-ready format with paired RGB, calibrated cameras, Pl\"ucker rays, and per-frame depth, normal, world-position, and tracking maps, so that alternative conditioning representations can be compared under matched renderer inputs.  Appendix~\ref{app:data_landscape} details the dataset landscape.

%% file: tog26_sections/03_method.tex
\label{sec:method}

\begin{figure*}[t]
  \centering
  \includegraphics[width=\textwidth]{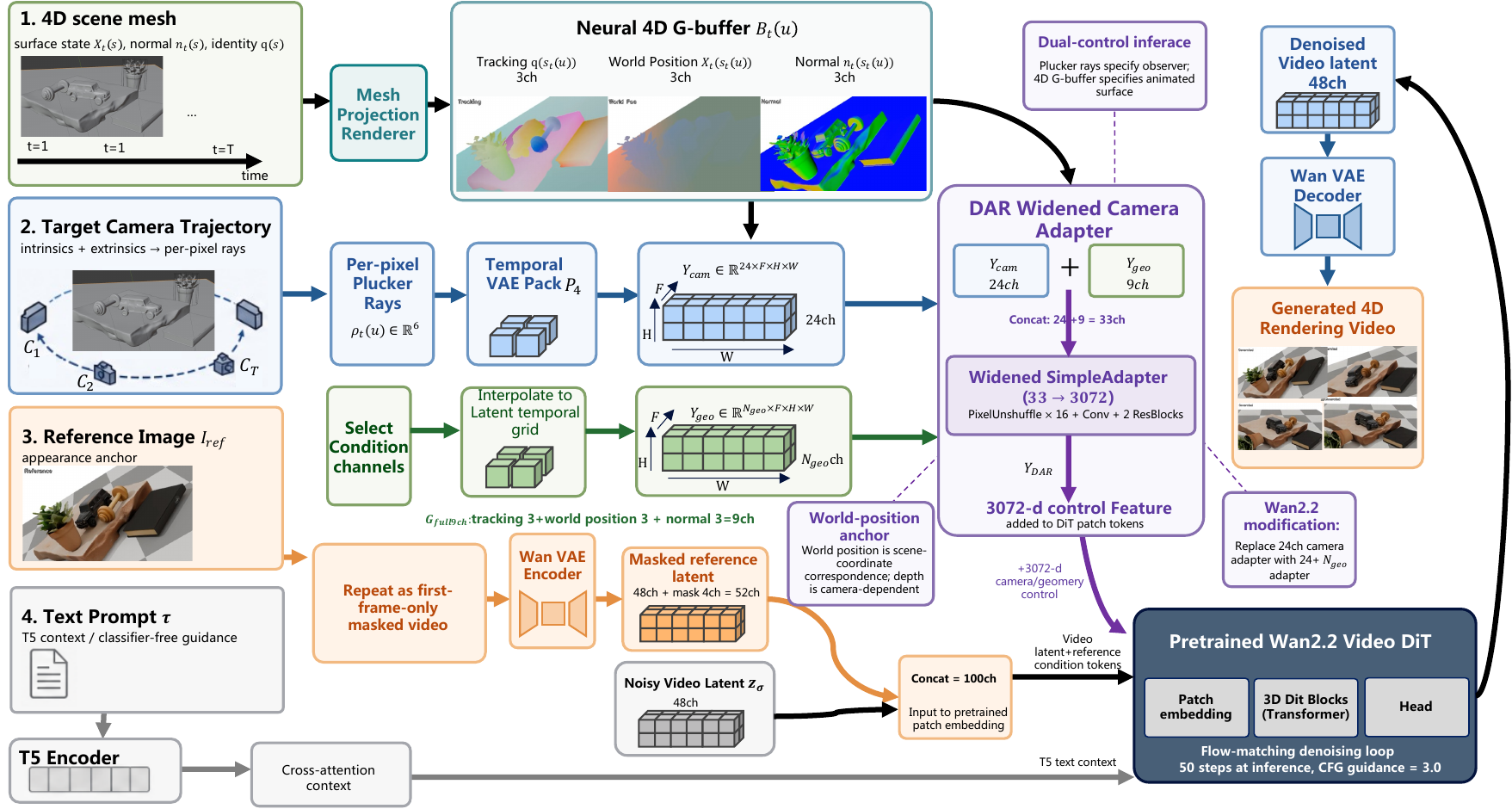}
  \caption{\textbf{DAR pipeline.} An animated 4D mesh, target camera trajectory, reference image, and text prompt are converted into aligned video-diffusion conditions.  Pl\"ucker rays encode the observer; a mesh projection renderer produces tracking, world-position, and normal maps; a widened SimpleAdapter injects the combined camera-plus-geometry control into the pretrained Wan2.2 video DiT.}
  \Description{Pipeline figure for DAR. Inputs are animated mesh, camera trajectory, reference image, and text prompt. The mesh is projected into tracking, world position, and normal buffers; camera rays are packed as Plucker channels; the combined condition is injected through a widened adapter into a pretrained Wan2.2 video diffusion transformer.}
  \label{fig:pipeline}
\end{figure*}

DAR is a representation-level extension of Wan2.2 camera control (Fig.~\ref{fig:pipeline}).  We keep the pretrained video diffusion substrate and Pl\"ucker-ray camera interface, but replace the camera-only control tensor by a joint camera-plus-4D-geometry tensor.  This section defines the rendering problem, the adapter change, and the role of tracking+world position as the visible 4D state code.

\subsection{Problem Formulation}\label{sec:method_problem}

Let $\mathcal{S}$ denote the canonical surface domain of an animated mesh.  At video time $t$, the animation maps a surface point $s\!\in\!\mathcal{S}$ to world position and normal
\begin{equation}
  X_t(s)\in\mathbb{R}^3,\qquad n_t(s)\in\mathbb{S}^2 .
\end{equation}
A camera $C_t=(K_t,R_t,o_t)$ projects visible surface points to pixels.  For pixel $u=(x,y)$, let $s_t(u)$ be the visible surface point selected by rasterization, if any.  The target renderer should produce
\begin{equation}
  \hat{\mathcal{V}}
  =
  G_\theta\!\left(
    I_{\rm ref},\,
    \{C_t\}_{t=1}^{T},\,
    \{X_t,n_t,s_t\}_{t=1}^{T},\,
    \tau
  \right),
  \label{eq:dar_task}
\end{equation}
where $I_{\rm ref}$ is the first-frame appearance reference and $\tau$ is an optional text prompt.  The geometry and camera are not latent variables to be invented by the model; they are user-specified state.

\subsection{From Wan2.2 Camera Control to DAR Control}\label{sec:method_wan_extension}

For each pixel $u$ and time $t$, Wan2.2-Control-Camera forms a Pl\"ucker ray
\begin{equation}
  \rho_t(u)=\big(d_t(u),\, o_t\times d_t(u)\big)\in\mathbb{R}^{6},
  \label{eq:plucker}
\end{equation}
where $d_t(u)$ is the world-space ray direction and $o_t$ is the camera center.  Wan2.2 temporally packs four neighboring frames to match the video VAE latent grid:
\begin{equation}
  Y_{\rm cam}=\mathcal{P}_{4}\!\left(\{\rho_t\}_{t=1}^{T}\right)
  \in \mathbb{R}^{24\times F\times H\times W}.
  \label{eq:camera_pack}
\end{equation}
The original control adapter maps $Y_{\rm cam}$ to transformer-width features and adds them to the DiT patch tokens.  This interface controls camera motion, but it says nothing about which animated surface lies on each ray.

DAR keeps the representation in Eq.~\eqref{eq:plucker}, the temporal packing in Eq.~\eqref{eq:camera_pack}, and the same residual injection point.  The change is the control tensor:
\begin{equation}
  Y_{\rm DAR}
  =
  Y_{\rm cam}
  \oplus
  \mathcal{P}_{\rm vae}(B)
  \in \mathbb{R}^{(24+N)\times F\times H\times W},
  \label{eq:dar_control_tensor}
\end{equation}
where $B$ is a mesh-projected geometry buffer and $\mathcal{P}_{\rm vae}$ resamples it to the latent temporal grid.  We replace the camera adapter by
\begin{equation}
  x^{(0)} = {\rm Patch}(z_\sigma)+A_\phi(Y_{\rm DAR}).
  \label{eq:dar_adapter}
\end{equation}
The main configuration uses $N{=}9$.  The adapter is the same lightweight \texttt{SimpleAdapter} family used by Wan2.2 camera control: pixel-unshuffle, stride-2 convolution, and two residual blocks, outputting transformer-width features.  Camera and animated geometry are therefore fused before the DiT reasons over the video, rather than appended as late guidance.

\subsection{Neural 4D G-buffer}\label{sec:method_3dgrounded}

For each visible pixel $u$ at time $t$, DAR projects three mesh-derived signals:
\begin{equation}
  B_t(u)
  =
  \left[
    q(s_t(u)),\,
    \bar X_t(s_t(u)),\,
    \bar n_t(s_t(u))
  \right]
  \in \mathbb{R}^{9}.
  \label{eq:gbuffer}
\end{equation}
Here $q(s)\in[0,1]^3$ is a persistent tracking color or instance/part identity, $\bar X_t$ is the scene-normalized world position, and $\bar n_t$ is the normal mapped to image range.  Pixels without a visible mesh hit are filled by the background convention used in the projection renderer and are masked consistently across all channels.

The three components serve different roles.  Tracking gives a persistent identity cue for appearance transport.  World position gives the current metric 3D state of the visible surface.  Normal gives first-order local shape for shading and silhouette detail.  We call this tensor a neural 4D G-buffer because it plays the role of a classical G-buffer (an image-format intermediate representation for rendering) but contains only the geometry needed by a reference-guided video diffusion renderer.

\subsection{Why Tracking+World Position Is the 4D Code}\label{sec:method_why_wp}

Depth is easy to render and widely supported by video models.  With calibrated rays, an ideal geometric decoder can back-project depth into 3D; the issue is the inductive bias of the conditioning interface.  A 4D renderer needs the condition image to expose both \emph{which persistent surface element} is visible and \emph{where that element is in the target 3D state}.  These are different variables: appearance is attached to surface identity, while silhouette, occlusion, and camera-relative layout depend on current world state.

\paragraph{Definition 1 (visible 4D state code).}
For a visible mesh hit, $\chi_t(u)$ is a visible 4D state code if it determines the pair
\begin{equation}
  \big(q(s_t(u)),\, X_t(s_t(u))\big)
  \label{eq:visible_4d_state}
\end{equation}
up to the tracking granularity, and if this code is independent of the observing camera.  DAR uses $\chi_t(u)=(q(s_t(u)),\bar X_t(s_t(u)))$: tracking indexes the persistent surface/part identity, while normalized world position gives the current scene-coordinate state.

\paragraph{Proposition 1 (depth is a camera chart; DAR is a scene-state chart).}
For view-z depth $d_C(X)=e_3^\top R(X-o)$, a single depth value is non-injective over 3D points and changes when the same point is observed by a translated camera.  Tracking alone identifies what to propagate but not where it should be rendered; world position alone gives a state coordinate but not the persistent identity needed for appearance transport when parts repeat, cross, or occlude.  Their pair $(q,\bar X)$ determines both variables in Eq.~\eqref{eq:visible_4d_state} within the scene bounds, while Pl\"ucker rays separately encode the observer.  Appendix~\ref{app:proof} gives the derivation and motion-factorization view.

\paragraph{Consequence for reference-guided rendering.}
Let $a(q)$ denote appearance attached to a persistent surface identity.  With $(q,\bar X)$, the model can learn appearance transport conditioned on explicit identity and current 3D state.  With depth, it must recover $X$ by combining depth, pixel location, and camera pose, while still inferring identity from appearance history.  Our controlled depth-swap ablation tests the geometric slot: DAR and the matched depth variant keep tracking, normal, backbone, data, and optimization fixed, and replace only world position by depth.

\subsection{Appearance Branch and Training Objective}\label{sec:method_train}

The reference image follows the Wan2.2 image-to-video inpainting interface.  We place $I_{\rm ref}$ in the first frame of a video tensor, zero-fill the remaining frames, and provide a binary mask indicating which latent positions are known.  This appearance condition is concatenated with the noisy latent in the original Wan2.2 path; DAR changes only the control adapter in Eq.~\eqref{eq:dar_adapter}.

Let $c=(I_{\rm ref},Y_{\rm DAR},\tau)$ collect all conditions.  With flow-matching noise level $\sigma$, the model predicts the scheduler target $v_\sigma$:
\begin{equation}
  \hat v_\theta
  =
  f_\theta(z_\sigma,\sigma,c),
  \qquad
  \mathcal{L}(\theta)
  =
  \mathbb{E}
  \left[
    \left\|
      f_\theta(z_\sigma,\sigma,c)-v_\sigma
    \right\|_2^2
  \right].
  \label{eq:loss}
\end{equation}
For LoRA experiments, the pretrained Wan2.2 transformer is frozen except for rank-256 LoRA modules on attention/FFN projections and the new \texttt{SimpleAdapter}.  The full-fine-tune variant trains all transformer parameters plus the same adapter.  The 24-channel camera-only ablation is the same widened-interface code path with $N{=}0$.

%% file: tog26_sections/04a_dataset.tex
\label{sec:dataset}

To train and evaluate a 4D renderer, each sample must expose the animated scene state, target camera trajectory, and dense per-frame projections in a video-model format.  We therefore build \textbf{DAR-4D}, a rendering corpus that builds on the OVOW pipeline for constructing paired video and 4D-scene data~\citep{chen2026videoworldturningmonocular}, separating large source pools from the smaller verified splits used for quantitative claims.

\subsection{Sources and Evaluated Splits}\label{sec:dataset_sources}

Table~\ref{tab:dar4d_sources_splits} lists both raw assets and evaluated clips.  This avoids conflating available rendering material with benchmark cases that have complete RGB, camera, geometry-buffer, inference, and metric verification.

\begin{table}[t]
  \centering
  \footnotesize
  \caption{\textbf{DAR-4D sources and evaluated splits.} Source counts come from local manifests; evaluated splits list clips with complete rendering, inference, and metric verification.}
  \label{tab:dar4d_sources_splits}
  \begin{tabularx}{\linewidth}{@{}p{0.29\linewidth}p{0.21\linewidth}X@{}}
    \toprule
    Component & Scale & Use in DAR-4D \\
    \midrule
    BlenderKit scene blends & 4{,}192 & Complex Blender scenes rendered under randomized camera paths. \\
    Unreal Engine archives & 2{,}588 & Indoor/outdoor environments for cinematic trajectories and clutter. \\
    Textured mesh records & 10.48M total; 82{,}725 strict textured & Object pool for synthetic scene assembly and appearance diversity. \\
    Z-OO rigged assets & 2{,}448 assets; 74 categories & Articulated animal FBX/BVH motion for 4D dynamics. \\
    \midrule
    \texttt{4d\_vis} main benchmark & 17 scenes $\times$ 4 paths = 68 clips & Main quantitative table, ablations, and external baseline suite. \\
    OOD-34 probe & 17 unseen scenes $\times$ 2 paths = 34 clips & Generalization check for unseen assets and trajectories. \\
    Reference variants & 10 clips & Appearance-control test with fixed geometry and camera. \\
    \bottomrule
  \end{tabularx}
\end{table}

Compared with existing 3D/4D resources, DAR-4D is organized around the paired tuple needed here: first-frame reference, target camera, animated mesh state, and dense geometry buffers.  Appendix~\ref{app:data_landscape} contrasts this interface with prior datasets.

\subsection{Procedural 4D Scene Synthesis}\label{sec:dataset_synthesis}

The synthetic pipeline turns assets into plausible animated scenes through randomized but constrained assembly, complementing code-driven parametric generation of CAD and Blender scenes \citep{chen2026paircoderpairprogramminguniversal}.  For the mixed-asset path, each scene samples multiple textured meshes, one or more rigged Z-OO animals, an HDRI environment, and a trajectory seed.  Static meshes are scaled into a compact physical range, placed by AABB rejection sampling with a collision margin, and optionally given simple keyframed motion such as spin, tilt, bob, slide, sway, or self-orbit.  Rigged animals are imported with their FBX animation, scaled to a target size, looped through NLA strips, and attached to drive empties so root motion can be composed with the skeletal walk cycle.

Motion planning is collision-aware.  Initial placement avoids overlaps with existing static and animated objects.  Walking paths are sampled as smooth piecewise trajectories in the ground plane, densified with Catmull-Rom interpolation, and rejected if the moving AABB intersects static objects or another moving path at the same frame.  If no valid path is found, the asset is kept stationary rather than allowing interpenetration.  This conservative rule keeps training conditions physically plausible.

Camera trajectories are generated from archetypes rather than fixed templates.  The Blender path uses 16 trajectory families, including short/half/full orbits, orbit+dolly, top-down orbits, low arcs, pure dolly, lateral pan, rolling orbit, orbit+pan, and vertical arcs.  The UE path uses a related 14-family generator with orbit, corridor, dolly, top-spin, and oblique motions.  Each trajectory randomizes span, start yaw, pitch, roll, dolly, pan, and radius within conservative bounds.  A fit-camera step computes animated scene bounds over sampled frames and clamps camera distance so the subject remains in view.

\subsection{Unified DAR Format and Quality Control}\label{sec:dataset_format}

After rendering, every sample is converted into a unified DAR bundle.  A sample directory contains \texttt{video.mp4}, the first-frame \texttt{reference.png}, camera intrinsics/extrinsics, renderer metadata, and per-frame projection files.  Each projection stores five aligned signal families: \texttt{tracking} for dense object/part identity, \texttt{world\_pos} for normalized world coordinates, \texttt{normal} for surface orientation, \texttt{depth} for a view-dependent baseline signal, and \texttt{sparse\_tracking} for point-track style controls.

Quality control combines automatic filters with renderer-side constraints.  The scripts reject or quarantine samples with near-uniform RGB, depth, normal, world-position, or tracking statistics, filter unsafe camera paths, verify scene-level world-position normalization, and inspect tracking/color buffers for consistency.  These checks keep the reported benchmark conservative: the full corpus is designed to scale, but the paper reports only subsets with complete rendering, inference, and metric verification.

%% file: tog26_sections/04_setup.tex
\label{sec:experiments}

The central comparison is a controlled condition-form ablation: all trainable variants share the same backbone, training data, compute budget, and optimization recipe; the variable is the geometry channel set provided to the widened \texttt{SimpleAdapter}.

\subsection{Protocol and Compared Methods}\label{sec:setup_bench}

The main evaluation set is the 68-case \texttt{4d\_vis} benchmark: 17 base scenes and 4 camera trajectories per scene.  All methods receive the same first-frame reference image, target camera trajectory, and mesh projection.  Resolution is fixed to $480{\times}832$; sequence length follows the ground-truth clip and may be 49, 77, 81, 89, 93, 117, or 249 frames.  We additionally report OOD-34, with 17 unseen scenes and 2 novel trajectories per scene, and a 10-case reference-variant subset that fixes geometry/camera while changing the reference image.

\begingroup
\emergencystretch=2em
We compare four groups.  \textbf{Off-the-shelf Wan2.2} includes Wan2.2-Camera, conditioned on Pl\"ucker rays only, and Wan2.2-Depth, conditioned on ground-truth depth videos and the first-frame reference~\citep{wang2025wan}.  \textbf{DAR LoRA variants} include the full condition with tracking, world position, and normal channels, together with 11 ablations.  Single-channel rows isolate tracking, normal, world position, and depth; leave-one-out rows remove one channel family from the full design; depth-swap rows replace world position by depth in comparable channel sets.  The complete configuration list is in Appendix~\ref{app:ablation_settings}.  \textbf{Full fine-tune} reports DAR at checkpoint 4000 without the LoRA bottleneck.  \textbf{External baselines} include DAS and VideoFrom3D, executed on the same 68 cases~\citep{gu2025diffusionshader,kim2025videofrom3d}.
\par
\endgroup

DAS and VideoFrom3D have useful but mismatched interfaces: DAS is image/tracking-guided object-centric synthesis, while VideoFrom3D assumes sparse anchors and mostly static scene geometry.  Their output length, resolution, and camera-control targets therefore do not exactly match our animated-mesh renderer protocol.  We resize outputs and sample frames by normalized animation time before computing the same metrics, and mark these rows as diagnostic rather than part of the controlled ablation ranking.

\subsection{Metrics}\label{sec:setup_metrics}

We report PSNR, SSIM~\citep{wang2004ssim}, and LPIPS~\citep{zhang2018lpips} as primary frame-aligned rendering metrics against the corresponding Blender ground-truth frame.  We also report TempL1, the mean L1 distance between adjacent-frame differences, and RefL1, the mean per-frame L1 distance to the first-frame reference.  RefL1 is auxiliary: a low value can indicate appearance retention, but it can also indicate that a method stays too close to the first view rather than respecting novel camera motion.  We leave tracking IoU, ATE/RTE, FVD, and VBench-style scores to future work because they require reverse-estimated masks, cameras, or perceptual judgments that would introduce additional estimator error into the current frame-aligned protocol.

\subsection{Training and Inference}\label{sec:setup_training}

All LoRA settings start from Wan2.2-Fun-5B-Control-Camera.  The backbone is frozen except for rank-256 LoRA modules on attention QKV/O and FFN projections, while the new \texttt{SimpleAdapter} is trained.  We use AdamW, learning rate $1{\times}10^{-4}$, constant-with-warmup scheduling with 200 warmup steps, batch size 1, gradient accumulation 1, 10{,}000 training steps, checkpoints every 2{,}000 steps, bf16 mixed precision, and gradient checkpointing.  Inference uses guidance scale 3.0, 50 flow-matching denoising steps, and seed 42.  Each LoRA run uses either $3{\times}$ A800-80GB or $2{\times}$ H100-80GB and takes roughly 36 hours.

For the full fine-tune, we train selected condition sets without LoRA on GPUs 1--7 ($7{\times}$ A800-80GB) using FSDP Full-Shard.  The backbone and SimpleAdapter are trained together, for roughly 5.4B trainable parameters.  Hyperparameters match the LoRA recipe except for a lower learning rate, $2{\times}10^{-5}$.  Sec.~\ref{sec:results_gen} reports DAR checkpoint 4000 on all 68 \texttt{4d\_vis} cases; Appendix~\ref{app:additional_quant} reports the matched checkpoint-2000 full-fine-tune ablations.

%% file: tog26_sections/05_results.tex
\label{sec:results}

We evaluate DAR on the \texttt{4d\_vis} split of DAR-4D: 17 OVOW/UE assets~\citep{chen2026videoworldturningmonocular} and 4 trajectory variants per asset, for 68 videos.  The comparison includes two off-the-shelf Wan2.2 baselines, two executed external baselines, 11 LoRA geometry ablations, and a full-fine-tune preview.  All LoRA settings share the same Wan2.2-Fun-5B-Control-Camera backbone, 10{,}000-step training budget, rank-256 LoRA setting, and hyperparameters; only the geometry channels input to \texttt{SimpleAdapter} change.

\subsection{Main Quantitative Comparison}\label{sec:results_main}

Table~\ref{tab:main_comparison} gives the 68-case comparison.  All LoRA rows are reported at checkpoint 10{,}000.  Rows marked with $\dagger$ use normalized-time alignment because the external system emits a non-matching video length or resolution; we include them as diagnostic reference points, but exclude them from the controlled rank ordering.

\begin{table*}[t]
  \centering
  \caption{\textbf{Main quantitative comparison on the DAR-4D 68-scene benchmark.} $\uparrow$ means higher is better and $\downarrow$ means lower is better.  \textbf{Bold} marks the best value among frame-aligned Wan/DAR rows and \underline{underline} marks the second best.  All trainable ablation rows share backbone, data, compute, and training recipe; the variable is the geometry channel set input to \texttt{SimpleAdapter}.  $\dagger$ rows are executed external baselines sampled by normalized animation time, so they are diagnostic rather than part of the controlled ablation ranking.}
  \label{tab:main_comparison}
  \resizebox{\textwidth}{!}{%
  \begin{tabular}{l l | c c c | c c}
    \toprule
    & Method & PSNR$\uparrow$ & SSIM$\uparrow$ & LPIPS$\downarrow$ & TempL1$\downarrow$ & RefL1$\downarrow$ \\
    \midrule
    \multirow{2}{*}{Off-the-shelf}
      & Wan2.2-Camera (no FT)~\citep{wang2025wan}              & 13.51 & 0.521 & 0.419 & 7.96 & \underline{33.84} \\
      & Wan2.2-Depth (no FT)~\citep{wang2025wan}               & 21.68 & 0.835 & 0.175 & \underline{6.99} & \textbf{29.55} \\
    \midrule
    \multirow{2}{*}{External$\dagger$}
      & DAS~\citep{gu2025diffusionshader}                      & 13.44 & 0.528 & 0.464 & 2.59 & 45.55 \\
      & VideoFrom3D (67/68)~\citep{kim2025videofrom3d}          & 8.04 & 0.371 & 0.652 & 3.61 & 89.30 \\
    \midrule
    \multirow{5}{*}{Single-channel}
      & Camera rays only (Pl\"ucker)                            & 12.66 & 0.488 & 0.436 & 15.96 & 47.16 \\
      & Depth only (1ch)                                       & 20.93 & 0.842 & 0.161 & \textbf{6.03} & 43.15 \\
      & Normal only (3ch)                                      & \underline{22.16} & \underline{0.869} & \underline{0.150} & 9.90 & 42.34 \\
      & World position only (3ch)                              & 21.71 & 0.836 & 0.169 & 9.97 & 42.84 \\
      & Tracking only (3ch)                                    & 19.36 & 0.808 & 0.219 & 12.90 & 47.52 \\
    \midrule
    \multirow{6}{*}{Multi-channel}
      & Tracking + normal (6ch, no world position)              & 18.74 & 0.777 & 0.220 & 12.97 & 35.83 \\
      & Tracking + world position (6ch, no normal)              & 18.94 & 0.789 & 0.211 & 12.57 & 35.21 \\
      & Normal + world position (6ch, no tracking)              & 19.73 & 0.802 & 0.210 & 12.68 & 34.32 \\
      & Tracking + depth (4ch)                                  & 20.94 & 0.836 & 0.169 & 10.15 & 43.36 \\
      & Normal + depth (4ch)                                    & 21.26 & 0.840 & 0.170 & 9.28 & 41.76 \\
      & Tracking + normal + depth (7ch) \emph{[world position $\to$ depth]} & 21.80 & 0.861 & 0.157 & 10.12 & 43.13 \\
    \midrule
    \rowcolor{gray!15}
    & \textbf{DAR (tracking + world position + normal)}       & \textbf{23.22} & \textbf{0.895} & \textbf{0.134} & 9.10 & 38.77 \\
    \bottomrule
  \end{tabular}%
  }
\end{table*}

\paragraph{DAR improves frame-aligned rendering quality.}
DAR reaches PSNR \textbf{23.22}, SSIM \textbf{0.895}, and LPIPS \textbf{0.134}, improving over off-the-shelf Wan2.2-Depth by \textbf{+1.54 dB} PSNR, \textbf{+0.060} SSIM, and \textbf{$-$0.041} LPIPS.  This is a strong comparison because Wan2.2-Depth receives an oracle ground-truth depth video at evaluation time.  The result indicates that the condition form matters: a joint Pl\"ucker plus 9-channel geometry interface is a better fit to our 4D rendering protocol than a single view-dependent depth signal.

\paragraph{World position beats depth when identity and shape are fixed.}
DAR and the matched depth-swap variant form the key controlled pair: same backbone, training recipe, and tracking/normal inputs, with only world position replaced by depth.  The world-position condition improves PSNR by \textbf{+1.42 dB}, SSIM by \textbf{+0.034}, and LPIPS by \textbf{$-$0.023} at checkpoint 10k, and remains ahead by \textbf{+1.26--+1.55 dB} PSNR across all saved checkpoints.  This supports Sec.~\ref{sec:method_why_wp}: when identity $q$ and local shape are exposed, the geometric state slot benefits from camera-independent world position rather than camera-dependent depth.

\paragraph{The geometry channels are complementary.}
Compared with the strongest single-channel LoRA variant, normal-only conditioning at 22.16 PSNR, DAR gains \textbf{+1.06 dB}.  Removing tracking, world position, or normal from the full design drops PSNR by 3.49, 4.48, and 4.28 dB respectively.  The 9-channel buffer is therefore not just channel expansion; tracking, world position, and normal carry different parts of the renderer state.

Together, these ablations match the prediction of Sec.~\ref{sec:method_why_wp}: tracking, world position, and normal are complementary.  The theory does not require world position alone to dominate every low-capacity LoRA row; it predicts that the full visible-state code should be strongest when identity and local shape are controlled, which is exactly what the matched depth-swap pair tests.

\subsection{Qualitative Results and Failure Modes}\label{sec:results_qual}

\begin{figure*}[t]
  \centering
  \makebox[\textwidth][c]{\includegraphics[width=1\textwidth]{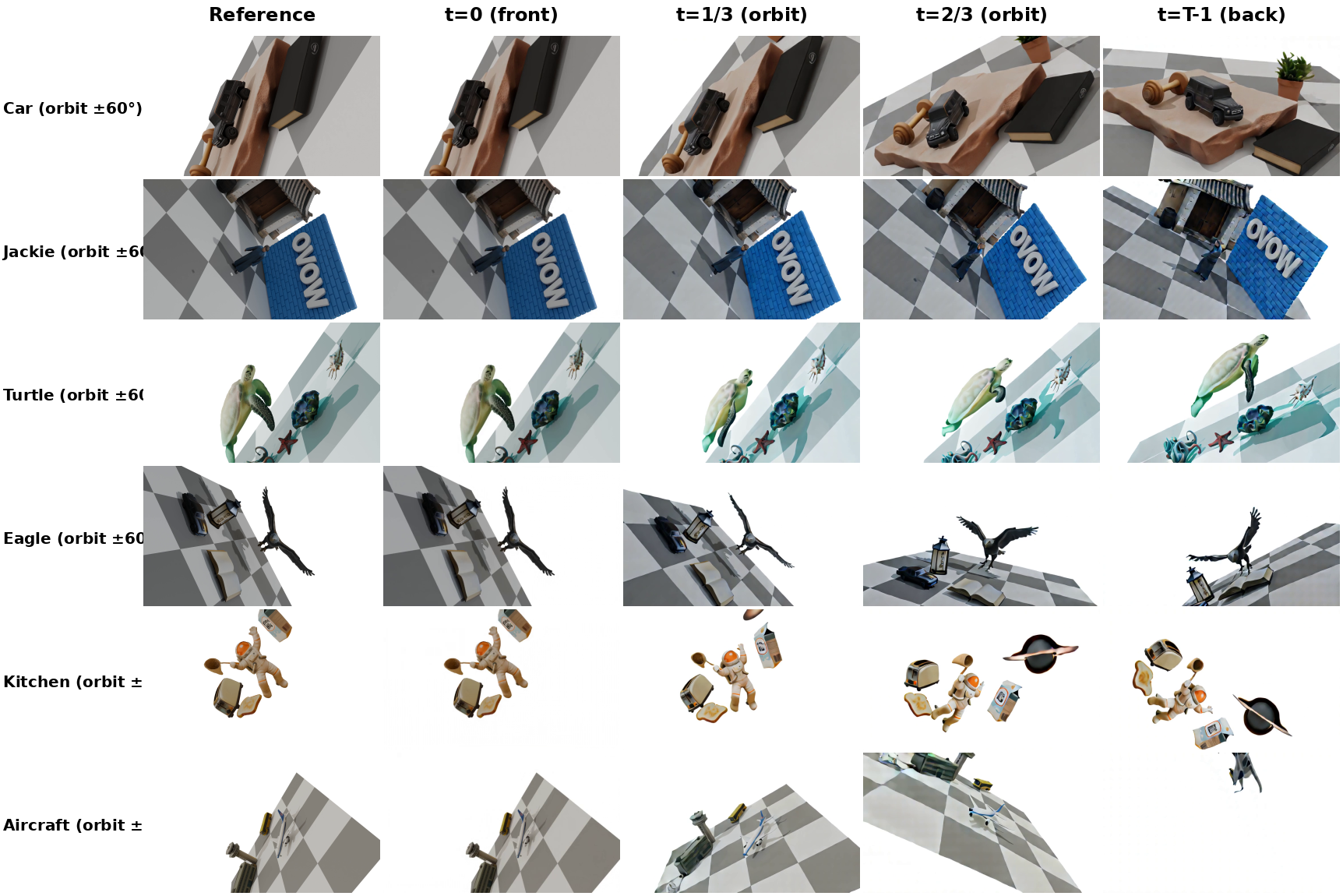}}
  \caption{\textbf{4D rotation gallery.} DAR outputs under a horizontal orbit ($\pm 60^{\circ}$) around six Blender meshes.  Each row shows one scene; columns show the first-frame reference and DAR outputs at $t{=}0,\,\tfrac{T}{3},\,\tfrac{2T}{3},\,T{-}1$.  DAR follows the target camera trajectory while preserving the animated mesh state and reference appearance across humans, animals, vehicles, mechanical objects, and indoor scenes.}
  \Description{4D rotation gallery of DAR outputs on six representative Blender scenes under a horizontal orbit camera trajectory.}
  \label{fig:rotation_gallery}
\end{figure*}

The post-reference figure pages show representative outputs.  Fig.~\ref{fig:main_qualitative} shows the final-frame comparison against Wan2.2-Camera, Wan2.2-Depth, and Blender GT under a $\pm60^{\circ}$ orbit, where the target view is farthest from the reference: Wan2.2-Camera hallucinates the foreground and Wan2.2-Depth couples camera and object motion, whereas DAR matches both the target camera and the animated mesh state.  Fig.~\ref{fig:baseline_suite} compares Wan2.2-Camera, Wan2.2-Depth, DAS, VideoFrom3D, and DAR at matched normalized animation times.  Camera-only control follows viewpoint but hallucinates the foreground; depth preserves coarse layout but entangles camera and object motion; DAS preserves some local identity but is object-centric; VideoFrom3D can synthesize plausible novel views but is tied to a mostly static-scene assumption.  DAR better follows both the target camera and the animated mesh state.  Fig.~\ref{fig:rotation_gallery} shows that the same interface handles humans, animals, vehicles, mechanical objects, and indoor scenes under large horizontal orbits.

\subsection{Generalization, Appearance Variants, and Full Fine-Tune}\label{sec:results_gen}

Appendix~\ref{app:additional_quant} adds four stress tests beyond Table~\ref{tab:main_comparison}.  First, trajectory-stratified results show that DAR is best on all four camera-motion groups, including the largest $\pm60^{\circ}$ orbit.  Second, the matched depth-swap comparison favors world position at every LoRA checkpoint from 2k to 10k.  Third, replacing persistent tracking by a binary foreground mask at checkpoint 2000 drops PSNR by 1.60 dB under LoRA and 1.27 dB under full fine-tuning.  Fourth, OOD-34 remains a stress test: Wan2.2-Depth reaches 22.25 PSNR, tracking+world-position conditioning reaches 20.99, and DAR reaches 19.04, indicating that the full 9-channel LoRA model still needs more capacity or data for open-category generalization.

The reference-variant subset tests appearance control by fixing one animated car mesh and camera while changing only the reference image, as shown in Fig.~\ref{fig:ref_variants}.  DAR changes paint style, color blocks, and vehicle appearance while keeping the camera, pose, and silhouette aligned, indicating partial decoupling between geometry and appearance.  To estimate the ceiling of DAR without the LoRA bottleneck, we also fully fine-tune the model.  Checkpoint 4000 reaches PSNR \textbf{25.36}, SSIM \textbf{0.917}, LPIPS \textbf{0.130}, and TempL1 \textbf{4.72} on all 68 cases: \textbf{+2.14 dB} above LoRA DAR ck10000 and \textbf{+3.68 dB} above off-the-shelf Wan2.2-Depth.

%% file: tog26_sections/07_conclusion.tex
We introduced 4D generative rendering: given an animated mesh scene, a target camera path, and a reference image, a video diffusion model should render the specified 4D state rather than invent a new one.  DAR builds on Wan2.2's Pl\"ucker-ray camera interface and adds a mesh-projected neural 4D G-buffer (tracking, world position, and normals) through the early control path, where tracking+world position serves as a camera-invariant state code that helps disentangle camera motion from animated surface motion and improves adherence to both trajectory and mesh state in our evaluations.   More broadly, our results suggest that the \,\emph{representation} of visible 4D state is a central ingredient for controllable 4D generative rendering: exposing persistent surface identity (tracking) together with scene-coordinate state (world position) provides a more faithful control signal than camera-dependent depth when both camera and object motion vary.

DAR is intentionally scoped as a renderer (not geometry reconstruction, mesh generation, or editable PBR material authoring) and assumes accurate 4D inputs plus a first-frame reference image; applying it to real reconstructions will likely require confidence masks and/or robustness to noisy geometry.  Generalization is still limited by finite synthetic supervision (68 clips; OOD-34 suggests remaining gaps in LoRA capacity and data diversity), motivating broader categories, richer motion/clutter/material coverage, and per-category reporting.  Finally, DAR is not real-time (480$\times$832, 81 frames, 50 steps: \mbox{\raisebox{0.25ex}{\textasciitilde}}1.5--2.5 minutes per clip on a single H100/A100), and videos beyond 200 frames can flicker under large camera changes and long occlusions, suggesting temporal condition smoothing and clip-level memory as next steps.

%% file: tog26_sections/08_appendix.tex
\section{Supplemental Details}

\subsection{Proof Details}\label{app:proof}

We restate the tracking/world-position argument from Sec.~\ref{sec:method_why_wp}.  Assume the tracking code $q$ is unique at the selected surface/part granularity and the scene normalization for $\bar X$ is injective inside the scene bounds.  Then $(q(s_t(u)),\bar X_t(s_t(u)))$ determines both the persistent visible identity and the current world-space state of the rasterized hit.  Dropping $q$ loses the identity variable needed for reference appearance transport; dropping $\bar X$ loses the metric target state needed for viewpoint, occlusion, and silhouette control.  Normal is not required for this identifiability statement, but supplies local differential shape.

For depth, fix a camera $C=(K,R,o)$ and choose any nonzero vector $v$ satisfying $e_3^\top Rv=0$.  Then $X$ and $X+v$ have the same view-z depth because
\begin{equation}
  d_C(X+v)=e_3^\top R(X-o)+e_3^\top Rv=d_C(X),
\end{equation}
but they are different 3D points.  Thus a scalar depth value cannot identify a surface point.  For camera dependence, translate the camera center by $\Delta=\alpha R^\top e_3$ while keeping orientation fixed.  The new depth is
\begin{equation}
  d_{C'}(X)=e_3^\top R(X-o-\Delta)=d_C(X)-\alpha,
\end{equation}
so the same world point receives a different condition value whenever $\alpha\neq0$.  By definition, normalized world position $w(X)=\bar X$ does not use $C$; after a fixed scene normalization, $\bar X=\bar X'$ implies $X=X'$ inside the scene bounding box.

The calibrated back-projection caveat is real.  If a pixel ray is $r_C(u)=R^\top K^{-1}\tilde u$, then
\begin{equation}
  X=o+
  {d_C(X)\over e_3^\top Rr_C(u)}\,r_C(u)
  \label{eq:depth_backproject}
\end{equation}
recovers world position from depth plus camera.  The representation issue is therefore not information loss under ideal geometry, but inductive bias: depth asks the video model to learn camera-dependent inverse projection while also learning appearance transport, whereas world position exposes the scene-coordinate correspondence directly.

\paragraph{Motion factorization.}
For a moving surface point,
\begin{equation}
  {d\over dt}d_{C_t}(X_t)
  =
  e_3^\top \dot R_t(X_t-o_t)
  +
  e_3^\top R_t(\dot X_t-\dot o_t),
  \label{eq:depth_motion_mix}
\end{equation}
so depth mixes object motion, camera translation, and camera rotation.  World position exposes $\dot X_t$ in the scene frame, while Pl\"ucker rays expose the camera.  DAR therefore factors the two user controls before they enter the DiT.

\paragraph{Identifiability corollary.}
Suppose two visible states $(q_1,X_1)\neq(q_2,X_2)$ receive the same condition value under a non-injective code $\kappa$.  Any renderer that conditions only on $\kappa$ and the reference appearance cannot distinguish the two states without additional context; appearance transport and target-state rendering are not identifiable from the code alone.  The pair $(q,\bar X)$ avoids this ambiguity at the chosen tracking granularity inside the scene bounds, while scalar depth avoids it only after being composed with pixel coordinates and camera pose and still does not explicitly expose persistent identity.

\subsection{Existing 4D / Dynamic-Scene Data}\label{app:data_landscape}

Table~\ref{tab:existing_4d_data} summarizes the closest data resources.  The key difference is the supervision tuple needed by DAR: first-frame reference, explicit target camera path, animated mesh state, and per-frame geometry buffers aligned to the generated RGB frames.

\begin{table}[h]
  \centering
  \scriptsize
  \caption{\textbf{Position of DAR-4D among existing 3D/4D data resources.} Existing resources are valuable, but most are organized for static 3D assets, dynamic scene understanding, stereo/depth estimation, or physics reasoning rather than reference-guided 4D generative rendering.}
  \label{tab:existing_4d_data}
  \begingroup
  \setlength{\tabcolsep}{3pt}
  \begin{tabularx}{\linewidth}{@{}>{\raggedright\arraybackslash}p{0.22\linewidth}>{\raggedright\arraybackslash}p{0.24\linewidth}>{\raggedright\arraybackslash}X@{}}
    \toprule
    Resource & Primary content & Gap for DAR-style rendering \\
    \midrule
    Objaverse-XL~\citep{deitke2023objaversexl} & 10M+ static 3D assets & Scale, but no native animation, target cameras, or condition videos. \\
    Kubric / MOVi~\citep{greff2022kubric} & Procedural dynamic videos with rich annotations & Excellent synthetic control, but not packaged as reference+\allowbreak mesh+\allowbreak camera renderer tuples. \\
    4DComplete~\citep{li2021deformingthings4d} & Non-rigid 4D object motion & Focuses on deforming shapes, not reference-guided scene video rendering. \\
    Dynamic Replica~\citep{karaev2023dynamicstereo} & Real dynamic stereo/depth sequences & Captures real motion, but mesh state and camera/\allowbreak appearance controls are not independently editable. \\
    DynamicVerse~\citep{dynamicverse} & Multimodal 4D world modeling data & Broad 4D understanding resource; not a matched ablation benchmark for mesh-derived renderer conditions. \\
    PhysInOne~\citep{physinone2026} & Visual physics learning and reasoning suite & Rich physical dynamics, but not organized around neural G-buffer conditioning for video diffusion renderers. \\
    \rowcolor{gray!12}
    DAR-4D & Animated meshes, cameras, RGB, depth, normal, world position, tracking & Built to test condition representations for 4D generative rendering. \\
    \bottomrule
  \end{tabularx}
  \endgroup
\end{table}

\subsection{Ablation Configurations}\label{app:ablation_settings}

\begin{table}[h]
  \centering
  \footnotesize
  \caption{\textbf{DAR and 11 LoRA ablations on DAR-4D.} All variants share the Wan2.2-Fun-5B backbone, SimpleAdapter + LoRA rank-256 training, the same 10k-step budget, and the same hyperparameters.  The only variable is the geometry channel set input to \texttt{SimpleAdapter}.}
  \label{tab:ablation_settings}
  \begin{tabularx}{\linewidth}{@{}lXc@{}}
    \toprule
    Condition set & Geometry channels & Ch. \\
    \midrule
    Camera rays only & none beyond Pl\"ucker rays                         & 24+0=24 \\
    Depth only       & depth (1)                                          & 24+1=25 \\
    Tracking only    & tracking (3)                                       & 24+3=27 \\
    Normal only      & normal (3)                                         & 24+3=27 \\
    World position only & world position (3)                              & 24+3=27 \\
    \midrule
    Tracking + normal & tracking + normal (6, no world position)           & 24+6=30 \\
    Tracking + world position & tracking + world position (6, no normal)    & 24+6=30 \\
    Normal + world position & normal + world position (6, no tracking)     & 24+6=30 \\
    Tracking + depth & tracking + depth (4)                               & 24+4=28 \\
    Normal + depth   & normal + depth (4)                                 & 24+4=28 \\
    Tracking + normal + depth & tracking + normal + depth (7)              & 24+7=31 \\
    \midrule
    DAR \textbf{[ours]} & tracking + world position + normal (9)          & 24+9=33 \\
    \bottomrule
  \end{tabularx}
\end{table}

\subsection{Additional Quantitative Results}\label{app:additional_quant}

Table~\ref{tab:main_comparison} is intentionally the controlled 10k-step LoRA ranking.  This appendix adds four diagnostic views of the same evidence: trajectory difficulty, checkpoint stability, explicit identity-code replacement, and OOD generalization.

\begin{table}[h]
  \centering
  \scriptsize
  \caption{\textbf{Trajectory-stratified results on the 68-case \texttt{4d\_vis} benchmark.} Each cell reports PSNR/SSIM/LPIPS.  DAR is best in all trajectory groups, including the largest $\pm60^{\circ}$ orbit.}
  \label{tab:trajectory_stress}
  \setlength{\tabcolsep}{2pt}
  \begin{tabular}{l c c c c}
    \toprule
    Trajectory & $n$ & Wan2.2-Depth & Depth swap & DAR \\
    \midrule
    dolly $\pm15^\circ$ & 17 & 22.49/0.864/0.151 & 22.63/0.892/0.133 & \textbf{22.96}/\textbf{0.910}/\textbf{0.120} \\
    orbit $\pm30^\circ$ & 17 & 21.94/0.837/0.179 & 21.64/0.853/0.165 & \textbf{25.61}/\textbf{0.921}/\textbf{0.124} \\
    orbit $\pm60^\circ$ & 17 & 20.60/0.789/0.209 & 20.89/0.824/0.185 & \textbf{21.56}/\textbf{0.851}/\textbf{0.163} \\
    orbit $0\to45^\circ$ & 17 & 21.69/0.852/0.162 & 22.03/0.876/0.146 & \textbf{22.74}/\textbf{0.897}/\textbf{0.131} \\
    \bottomrule
  \end{tabular}
\end{table}

\begin{table}[h]
  \centering
  \scriptsize
  \caption{\textbf{LoRA checkpoint sweep for the matched world-position/depth swap.} DAR keeps tracking and normal fixed and uses world position; the depth-swap variant replaces world position with depth.  The PSNR advantage is positive at every saved checkpoint.}
  \label{tab:lora_depth_swap_curve}
  \setlength{\tabcolsep}{3pt}
  \begin{tabular}{c c c c c c c c}
    \toprule
    Step & \multicolumn{3}{c}{DAR} & \multicolumn{3}{c}{Depth swap} & $\Delta$PSNR \\
    & PSNR & SSIM & LPIPS & PSNR & SSIM & LPIPS & DAR$-$swap \\
    \midrule
    2000  & 21.41 & 0.828 & 0.182 & 19.86 & 0.770 & 0.225 & +1.55 \\
    4000  & 22.57 & 0.863 & 0.158 & 21.31 & 0.818 & 0.187 & +1.26 \\
    6000  & 22.25 & 0.863 & 0.156 & 20.96 & 0.827 & 0.179 & +1.29 \\
    8000  & 22.88 & 0.888 & 0.140 & 21.58 & 0.853 & 0.159 & +1.30 \\
    10000 & 23.22 & 0.895 & 0.134 & 21.80 & 0.861 & 0.157 & +1.42 \\
    \bottomrule
  \end{tabular}
\end{table}

\begin{table}[h]
  \centering
  \scriptsize
  \caption{\textbf{Tracking identity and world-position stress test at checkpoint 2000.} Replacing persistent tracking by a binary mask weakens rendering even when world position and normal remain available.  The pattern holds both with LoRA and with full fine-tuning.}
  \label{tab:tracking_mask_depth}
  \setlength{\tabcolsep}{3pt}
  \begin{tabularx}{\linewidth}{l X c c c c}
    \toprule
    Training & Condition & PSNR & SSIM & LPIPS & TempL1 \\
    \midrule
    LoRA & tracking + world position + normal & \textbf{21.41} & \textbf{0.828} & \textbf{0.182} & 10.10 \\
    LoRA & binary mask + world position + normal & 19.81 & 0.767 & 0.223 & 6.83 \\
    LoRA & binary mask + depth + normal & 19.97 & 0.793 & 0.207 & \textbf{5.90} \\
    \midrule
    Full-FT & tracking + world position + normal & \textbf{23.27} & \textbf{0.877} & \textbf{0.153} & \textbf{4.58} \\
    Full-FT & binary mask + world position + normal & 22.00 & 0.849 & 0.164 & 4.71 \\
    Full-FT & binary mask + depth + normal & 19.66 & 0.797 & 0.208 & 4.64 \\
    \bottomrule
  \end{tabularx}
\end{table}

\begin{table}[h]
  \centering
  \scriptsize
  \caption{\textbf{Selected OOD-34 stress-test results.} OOD-34 uses unseen scenes and novel trajectories.  We report it as a generalization diagnosis, not as the primary controlled representation ranking.}
  \label{tab:ood34}
  \setlength{\tabcolsep}{3pt}
  \begin{tabular}{l c c c c c}
    \toprule
    Method & PSNR & SSIM & LPIPS & TempL1 & RefL1 \\
    \midrule
    Wan2.2-Camera & 10.44 & 0.538 & 0.466 & 9.44 & 74.53 \\
    Wan2.2-Depth & \textbf{22.25} & \textbf{0.913} & \textbf{0.118} & \textbf{2.46} & 27.01 \\
    Depth only & 20.98 & 0.837 & 0.161 & 4.19 & 25.74 \\
    Tracking only & 17.15 & 0.734 & 0.297 & 4.78 & 39.50 \\
    Tracking + world position & 20.99 & 0.845 & 0.149 & 3.86 & \textbf{23.50} \\
    Tracking + normal + depth & 20.16 & 0.819 & 0.174 & 4.29 & 23.92 \\
    DAR & 19.04 & 0.812 & 0.190 & 3.77 & 26.60 \\
    \bottomrule
  \end{tabular}
\end{table}

The OOD result does not contradict the main claim.  On the matched 68-case benchmark, the depth-swap comparison keeps all non-position channels fixed and favors world position at every LoRA checkpoint.  On OOD-34, the lighter tracking+world-position variant can outperform DAR, suggesting that rank-256 LoRA capacity and current training diversity are limiting open-category generalization for the full 9-channel condition.

\subsection{User Study and Additional Stress Tests}\label{app:user_study}

The current quantitative evidence is frame-aligned and controlled, which is appropriate for a renderer.  A user study can complement it by asking whether viewers perceive the intended controls.  We recommend a two-alternative forced-choice study with three questions per pair: (1) which video better follows the target camera path, (2) which better follows the animated object motion, and (3) which better preserves the first-frame appearance.  Each trial should show the first-frame reference, a compact visualization of the target condition, and two anonymized videos sampled from the same case.  The main pairings are DAR vs. Wan2.2-Depth, DAR vs. Wan2.2-Camera, and DAR vs. the matched depth-swap variant.  The last pairing is the most important because it isolates world position from depth while keeping tracking and normal fixed.

For a submission version with completed human results, report the preference rate, 95\% bootstrap confidence interval, and binomial test against 50\% for each question and pairing.  Stratify by camera trajectory severity and by foreground category if the number of trials permits.  We do not insert synthetic preference numbers here; the protocol is included to make the next evidence collection reproducible.

Two additional automatic stress tests would further support the method claim.  First, a \emph{correspondence stress test} should evaluate large-orbit cases where first and last views have minimal overlap, reporting endpoint PSNR/LPIPS and a mask-restricted foreground metric.  Second, a \emph{motion-separation stress test} should pair the same object animation with multiple camera paths and the same camera path with multiple object animations.  The expected failure mode is that depth-only control degrades when camera and object motion are recombined, while world-position conditioning remains stable because the camera and surface-state factors stay separated.